\documentclass[runningheads, dvipsnames]{llncs}
\pdfoutput=1
\usepackage{makeidx}
\usepackage{graphicx}
\usepackage{amsmath,amssymb} 
\usepackage{color}

\usepackage{epsfig}
\usepackage{graphicx}
\usepackage{bmpsize}
\usepackage{layouts}
\usepackage{amsmath}
\usepackage{amssymb}

\usepackage[utf8]{inputenc} 
\usepackage[T1]{fontenc}    
\usepackage{url}            
\usepackage{booktabs}       
\usepackage{colortbl}       
\usepackage{amsfonts}       
\usepackage{microtype}      
\usepackage{placeins}
\usepackage{mathtools}
\usepackage{dsfont}
\usepackage{todonotes}
\usepackage[ruled,vlined]{algorithm2e}
\usepackage{xcolor}
\usepackage{enumerate}
\usepackage[shortlabels]{enumitem}
\usepackage{caption}
\usepackage{subcaption}
\usepackage[export]{adjustbox}
\graphicspath{{./figures/}} 
\usepackage{wrapfig}

\usepackage{tikz}
\usetikzlibrary{arrows, arrows.meta, shapes, calc, positioning, decorations.pathreplacing, decorations.markings, fit, backgrounds}

\usepackage[pagebackref=true,breaklinks=true,letterpaper=true,colorlinks,bookmarks=false]{hyperref}
\usepackage[capitalize]{cleveref}

\newcommand{\x}{{x}}

\newcommand{\z}{{z}}
\newcommand{\cnet}{\varphi}
\newcommand{\cml}{\mathcal{L}_\mathrm{cML}}
\newcommand{\cond}{c}
\newcommand{\uu}{u}
\newcommand{\vv}{v}
\newcommand{\bolda}{{a}}
\newcommand{\boldb}{{b}}
\newcommand{\boldL}{{L}}

\DeclareMathSymbol{\sminus}{\mathbin}{AMSa}{"39}

\definecolor{vll-orange}{HTML}{E37238}
\definecolor{vll-green}{HTML}{96BF0D}
\definecolor{vll-dark}{HTML}{464646}
\definecolor{vll-light}{HTML}{757575}

\newcommand{\bftab}{\fontseries{b}\selectfont} 

\begin{document}
	\pagestyle{headings}
	\mainmatter

	\def\GCPR20SubNumber{70}

    \title{Conditional Invertible Neural Networks\\ for Diverse Image-to-Image Translation}

	\titlerunning{Conditional Invertible Neural Networks}
	\authorrunning{Ardizzone et al.}
	\author{Lynton Ardizzone, Jakob Kruse, Carsten L\"uth, Niels Bracher,\\Carsten Rother, Ullrich K\"othe}
	\institute{Visual Learning Lab, Heidelberg University}

\maketitle

\begin{abstract}
\noindent %
%
We introduce a new architecture called a conditional invertible neural network (cINN),
and use it to address the task of diverse image-to-image translation for natural images.
This is not easily possible with existing INN models due to some fundamental limitations.
%
The cINN combines the purely generative INN model with an unconstrained feed-forward network, 
which efficiently preprocesses the conditioning image into maximally informative features.
All parameters of a cINN are jointly optimized with a stable, maximum likelihood-based training procedure.
%
Even though INN-based models have received far less attention in the literature than GANs,
they have been shown to have some remarkable properties absent in GANs, e.g. apparent immunity to mode collapse.
%
We find that our cINNs leverage these properties for image-to-image translation,
demonstrated on day to night translation and image colorization.
Furthermore, we take advantage of our bidirectional cINN architecture to explore and manipulate emergent properties of the latent space, 
such as changing the image style in an intuitive way. \\
Code \& Appendix: \href{https://github.com/VLL-HD/conditional_INNs}{\tt \small github.com/VLL-HD/conditional\_INNs}\\
\vspace{-5mm}
\end{abstract}

\section{Introduction}
\vspace{-1mm}
INNs occupy a growing niche in the space of generative models.
Because they became relevant more recently compared to GANs or VAEs,
they have received much less research attention so far.
Currently, the task of image generation is still dominated by GAN-based models \cite{brock2018large,karras2017progressive,karras2019style}.
Nevertheless, INNs have some extremely attractive theoretical and practical properties, leading to an increased research interest recently:
The training is not adversarial, very stable, and does not require any special tricks.
Their loss function is quantitatively meaningful for comparing models, checking overfitting, etc. \cite{theis2015note}, which is not given with GANs.
INNs also do not experience the phenomenon of mode collapse observed in GAN-based models \cite{salimans2016improved}.
Compared to VAEs, they are able to generate higher-quality results, because no ELBO approximation or reconstruction loss is needed, 
which typically leads to modeling errors \cite{zhao2019infovae,alemi2018fixing}.
Furthermore, they allow mapping real images into the latent space for explainability, interactive editing, and concept discovery \cite{kingma2018glow,jacobsen2018excessive}.
In addition, they have various connections to information theory,
allowing them to be used for lossless compression \cite{hoogeboom2019integer}, information-theoretic training schemes \cite{ardizzone2020exact}, 
and principled out-of-distribution detection \cite{choi2018waic,nalisnick2019detecting}.

In this work, we present a new architecture called a conditional invertible neural network (cINN), and apply it to diverse image-to-image translation.
Diverse image-to-image translation is a particular conditional generation task:
given a conditioning image $Y$, the task is to model the conditional probability distribution $p(X|Y)$
over some images $X$ in a different domain.
`\emph{Diverse}' implies that the model should generate different $X$ covering the whole distribution, not just a single answer.
More specifically, we consider the case of paired training data, meaning matching pairs $(x_i, y_i)$ are given in the training set.
Unpaired image-to-image translation in theory is ill-posed, and only possible through inductive bias or explicit regularization.

For this setting, the use of existing INN-based models has so far not been possible in a general way.
Some methods for conditional generation using INNs exist,
but these are mostly class-conditional,
or other cases where the condition directly contains the necessary high-level information 
\cite{sorrenson2020disentanglement,ardizzone2020exact,prenger2019waveglow}.
This is due to the basic limitation that
each step in the network must be invertible.
Because the condition itself is not part of the invertible transformation,
it can therefore not be passed across layers.
As a result, it is impossible for an INN to extract useful high-level features from the condition.
That would be necessary for effectively performing diverse image-to-image translation,
where e.g. the semantic context of the condition is needed.

\begin{wrapfigure}{R}{0.5\linewidth}
    \vspace{-10mm}
    \includegraphics[width=\linewidth]{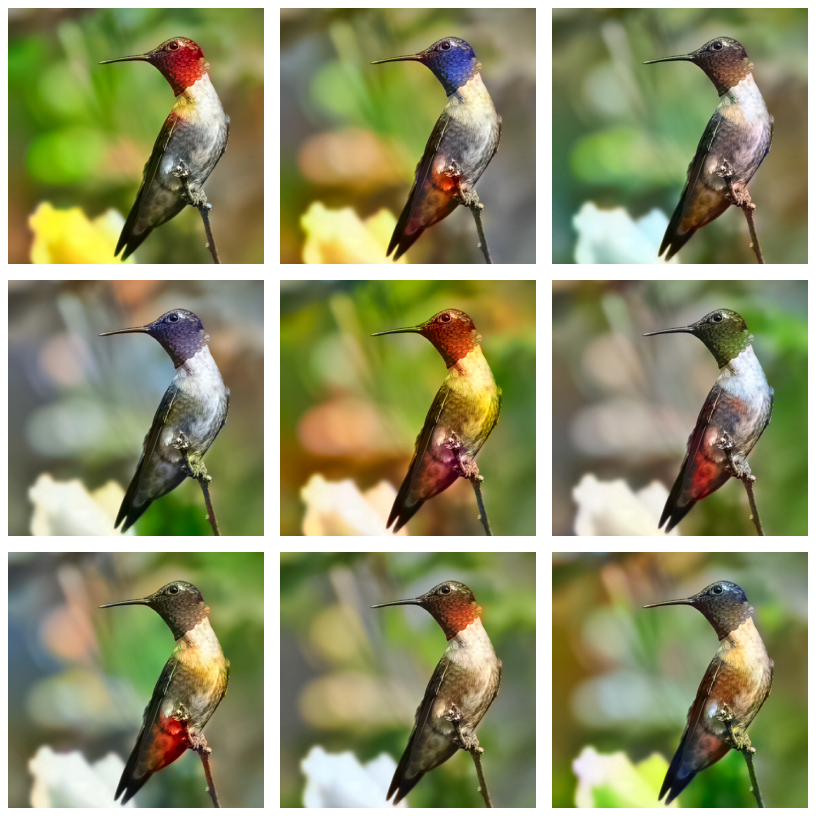}\vspace{-2mm}
	\caption{Diverse colorizations, which our network created for the same grayscale image. 
    One of them shows ground truth colors, but which?
    Solution at the bottom of next page. 
	}
	\vspace{-8mm}
	\label{fig:BIRD}
\end{wrapfigure}

\begin{figure}
    \parbox{0.64\linewidth}{
        \resizebox{\linewidth}{!}{\hspace{-40mm} \input{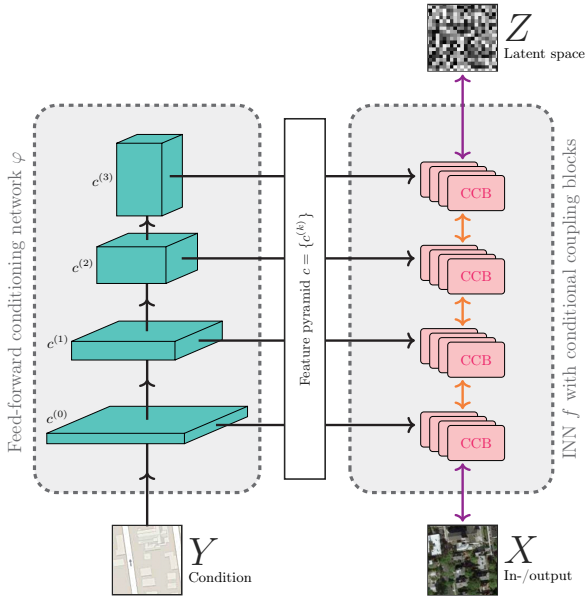}}
    } \hfill
    \parbox{0.35\linewidth}{
        \captionof{figure}{
        Illustration of the cINN. It consists of a feed-forward conditioning network (\emph{left half}), and an invertible part (\emph{right half}). \\
        \textbf{\textcolor{Black}{Black arrows:}} connections always in the same direction. \\
        \textbf{\textcolor{Emerald}{Green boxes:}} extracted feature maps $c^{(k)}$. \\
        \textbf{\textcolor{Plum}{Purple arrows:}} invertible connections, depending on training/testing.  \\
        \textbf{\textcolor{Orange}{Orange arrows:}} invertible wavelet downsampling.  \\
        \textbf{\textcolor{VioletRed}{Pink blocks:}} conditional coupling blocks (CCBs). 
        }
        \label{fig:overview}
    }
\vspace{-12mm}
\end{figure}

Our cINN extends standard INNs in three aspects to avoid this shortcoming.
Firstly, we use a simple but effective way to inject conditioning 
into the core building blocks at multiple resolutions in the form of so-called \emph{conditional coupling blocks} (CCBs).
Secondly, to provide useful conditions at each resolution level, we couple the INN with a feed-forward \emph{conditioning network}:
it produces a feature pyramid $C$ from the condition image $Y$, that can be injected into the CCBs at each resolution.
Lastly, we present a new invertible pooling scheme based on wavelets, that improves the generative capability of the INN model.
The entire cINN architecture is visualized in Fig.~\ref{fig:overview}.

The whole cINN can be trained end-to-end with a single maximum likelihood loss function,
leading to simple, repeatable, and stable training, without the need for hyperparameter tuning or special tricks.
We show that the learned conditioning features $C$ are maximally informative for the task at hand from an information theoretic standpoint.
We also show that the cINN will learn the true conditional probability if the networks are powerful enough.

Our contributions are summarized as follows:
\begin{itemize}
    \item We propose a new architecture called conditional invertible neural network (cINN), which combines an INN with an unconstrained feed-forward network for conditioning. 
    It generates diverse images with high realism, while adding noteworthy and useful properties compared to existing approaches.

    \item We demonstrate a stable, maximum likelihood training procedure for jointly optimizing the parameters of the INN and the conditioning network.  
    We show that our training causes the conditioning network to extract maximally informative features from the condition, measured by mutual information.
    
    \item We take advantage of our bidirectional cINN architecture to explore and manipulate emergent properties of the latent space. 
    We illustrate this for day-to-night image translation and image colorization. 
    \vspace{-6mm}
\end{itemize}

\tikz[remember picture, overlay] \node [rotate=180, anchor=south west, font=\footnotesize, text=black!50] at ($(current page.south east)+(-3.22cm,3.5)$) {Quiz solution: Bottom row, center image};

\section{Related work}
Image-to-image translation for natural images was first demonstrated with GAN-based models \cite{isola2017image}.
It was also extended to the unpaired setting by \cite{zhu2017cyclegan}.
However, these models are generally not able to produce diverse outputs.
Several works attempt to prevent such mode collapse in image-to-image GANs
through specialized architectures and regularization \cite{zhu2017toward,park2019semantic,lee2018diverse}.
A hybrid approach between GAN and autoencoder is used in \cite{ulyanov2018it} for diversity.
While these approaches do lead to visual diversity, there is currently no way to verify if they truly cover the entire distribution, or a lower-dimensional manifold.

Conditional INN models can be divided into methods with a conditional latent space, and methods where the INN itself is conditional.
Apart from our cINN, the only example for the second case to our knowledge is \cite{prenger2019waveglow}:
an INN-based model is used to de-modulate mel-spectrograms back into audio waves.
While the conditioning scheme is similar to our CCBs,
the condition is given externally and directly contains the needed information, instead of being learned. 
Diversity is also not considered, the model is only used to produce a single output for each condition.
For the second category of conditional latent space models, pixel-wise conditioning is in general more difficult to achieve.
\cite{kingma2018glow} manipulate latent space after training to generate images with certain global attributes.
In \cite{sorrenson2020disentanglement}, a class-conditional latent space is used for training to obtain a class-conditional INN model.
A special type of conditional latent space is demonstrated in \cite{ardizzone2018analyzing},
suitable for non-stochastic inverse problems of small dimensionality.
Examples where the approach is extended to spatial conditioning include
\cite{sun2019dual}, where two separate INNs define a mapping between medical imaging domains.
The model requires an additional loss term with hyperparameters, that has an unknown effect on the output distribution,
and diversity is not considered.
Closest to our work is \cite{kondo2019flow}, where a VAE and INN are trained jointly, to allow a specific form of diverse image-to-image translation.
However, the method is only applied for translation between images of the the same domain, 
i.e. generate similar images given a conditioning image.
The training scheme requires four losses that have to be balanced with hyperparameters.
Our cINN can map between arbitrary domains, is more flexible due to the CCB design instead of a conditional latent space, and only uses a single loss function to train all components jointly.

\section{Method}
\label{sec:methods}
\vspace{-2mm}
We divide this section into two parts:
First, we discuss the architecture itself,
split into the invertible components 
(Fig.~\ref{fig:overview} right),
and the feed-forward conditioning network
(Fig.~\ref{fig:overview} left).
Then, we present the training scheme and its effects on each component.
\vspace{-2mm}
\subsection{cINN Architecture}
\vspace{-0mm}
\noindent \textbf{Conditional coupling blocks.}
Our method to inject the conditioning features into the INN
is an extension of the affine coupling block architecture established by \cite{dinh2016density}.
There, each network block splits its input $\uu$ into two parts $[\uu_1, \uu_2]$ and applies affine transformations between them that have strictly upper or lower triangular Jacobians:
\begin{equation}
  \vv_1 = \uu_1 \odot \exp\big(s_1(\uu_2)\big) + t_1(\uu_2) \ , \quad 
  \vv_2 = \uu_2 \odot \exp\big(s_2(\vv_1)\big) + t_2(\vv_1) \ .
  \label{eq:unconditional_forward}
\end{equation}
The outputs $[\vv_1, \vv_2]$ are concatenated again and passed to the next coupling block.
The internal functions $s_j$ and $t_j$ can be represented by arbitrary neural networks, we call these the \emph{subnetworks} of the block.
In practice, each $[s_j, t_j]$-pair is jointly modeled by a single subnetwork, instead of separately.
Importantly, the subnetworks are only ever evaluated in the forward direction, even when the coupling block is inverted:
\begin{equation}
  \uu_2 = \big(\vv_2 - t_2(\vv_1) \big) \oslash \exp\big(s_2(\vv_1)\big) \ , \quad
  \uu_1 = \big(\vv_1 - t_1(\uu_2) \big) \oslash \exp\big(s_1(\uu_2)\big) \ .
  \label{eq:unconditional_inverse}
\end{equation}
As shown by \cite{dinh2016density}, the logarithm of the Jacobian determinant for such a coupling block 
is simply the sum of $s_1$ and $s_2$ over image dimensions, which we use later.

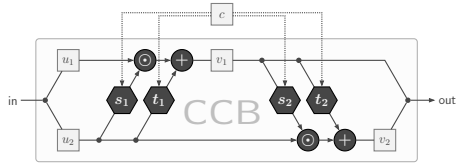
\begin{wrapfigure}{R}{0.5\linewidth}
    \vspace{-4mm}
	\resizebox{\linewidth}{!}{ \begin{tikzpicture}[
    every node/.style = {inner sep = 0pt, outer sep = 0pt, anchor = center, align = center, font = \sffamily\large, text = black!75},
    var/.style = {rectangle, minimum width = 2em, minimum height = 2em, text depth = 0, line width = 1pt, draw = black!50, fill = black!5},
    op/.style = {circle, minimum width = 2em, text depth = 2pt, line width = 1pt, draw = black, fill = vll-dark, text = black!5, font = \Large\boldmath},
    nn/.style = {op, regular polygon, regular polygon sides = 6, minimum width = 1cm},
    dot/.style = {circle, minimum width = 5pt, fill = vll-dark},
    connect/.style = {line width = 1pt, draw = vll-dark},
    arrow/.style = {connect, -{Triangle[length=6pt, width=4pt]}}]

    \node [anchor = east] (in) at (0,0) {in};

    \node [dot] (split) at ([shift = {(in.east)}] 0:0.9) {};
    \draw [connect, shorten < = 3pt] (in) to (split);

    \node [var] (u1) at ([shift = {(split)}] 60:1.5) {$\uu_1$};
    \draw [connect] (split) to (u1);

    \node [var] (u2) at ([shift = {(split)}] -60:1.5) {$\uu_2$};
    \draw [connect] (split) to (u2);

    \node [dot] (d1) at ([shift = {(u2)}] 0:1) {};
    \node [dot] (d2) at ([shift = {(d1)}] 0:1.2) {};

    \node [op] (mult2) at ([shift = {(d1)}] 60:3) {$\odot$};
    \draw [arrow] (d1) to (mult2);
    \draw [arrow] (u1) to (mult2);

    \node [op] (add2) at ([shift = {(d2)}] 60:3) {$+$};
    \draw [arrow] (d2) to (add2);
    \draw [arrow] (mult2) to (add2);

    \node [var] (v1) at ([shift = {(add2)}] 0:1.3) {$\vv_1$};
    \draw [connect] (add2) to (v1);

    \node [dot] (d3) at ([shift = {(v1)}] 0:1.3) {};
    \node [dot] (d4) at ([shift = {(d3)}] 0:1.2) {};

    \node [op] (mult1) at ([shift = {(d3)}] -60:3) {$\odot$};
    \draw [arrow] (d3) to (mult1);
    \draw [arrow] (u2) to (mult1);

    \node [op] (add1) at ([shift = {(d4)}] -60:3) {$+$};
    \draw [arrow] (d4) to (add1);
    \draw [arrow] (mult1) to (add1);

    \node [var] (v2) at ([shift = {(add1)}] 0:1.3) {$\vv_2$};
    \draw [connect] (add1) to (v2);

    \node [dot] (cat) at ([shift = {(v2)}] 60:1.5) {};
    \draw [connect] (v2) to (cat);
    \path [connect] (v1) -- (v1.center -| v2.center) -- (cat);

    \node [anchor = west] (out) at ([shift = {(cat)}] 0:1) {out};
    \draw [arrow, shorten > = 3pt] (cat) to (out);

    \node [nn] (s2) at ([shift = {(d1)}] 60:1.5) {$s_1\vphantom{t}$};
    \node [nn] (t2) at ([shift = {(d2)}] 60:1.5) {$t_1$};
    \node [nn] (s1) at ([shift = {(d3)}] -60:1.5) {$s_2\vphantom{t}$};
    \node [nn] (t1) at ([shift = {(d4)}] -60:1.5) {$t_2$};

    \node [var] (cnet) at ([shift = {(v1)}] 90:1.5) {$\cond\vphantom{t}$};
    \draw [arrow, densely dotted] ([shift = {(cnet.west)}] 90:0.05) -- ([shift = {(cnet -| s2.north)}] 90:0.05) -- (s2.north);
    \draw [arrow, densely dotted] ([shift = {(cnet.west)}] -90:0.05) -- ([shift = {(cnet -| t2.north)}] -90:0.05) -- (t2.north);
    \draw [arrow, densely dotted] ([shift = {(cnet.east)}] -90:0.05) -- ([shift = {(cnet -| s1.north)}] -90:0.05) -- (s1.north);
    \draw [arrow, densely dotted] ([shift = {(cnet.east)}] 90:0.05) -- ([shift = {(cnet -| t1.north)}] 90:0.05) -- (t1.north);

    \coordinate (top left) at ($(u1.north west) + (-2em, 1em)$);
    \coordinate (bottom right) at ($(v2.south east) + (2em, -1em)$);
    \begin{scope}[on background layer]
        \node [rounded corners = 3pt, fill = black!1, draw = black!33, fit = (top left) (bottom right)] (bg) {};
        \node [font = \Huge \bfseries \sffamily, color = black!25, scale=1.3] (CC) at ([shift={(v1)}] -90:1.7) {CCB};
    \end{scope}

\end{tikzpicture} }%
    \caption{A single conditional coupling block (CCB).}
    \label{fig:conditional-coupling}
    \vspace{-6mm}
\end{wrapfigure}

We adapt the design of \cref{eq:unconditional_forward,eq:unconditional_inverse} to produce a conditional coupling block (CCB):
Because the subnetworks $s_j$ and $t_j$ are never inverted, we can concatenate conditioning data $\cond$ to their inputs without losing the invertibility, 
replacing $s_1(\uu_2)$ with $s_1(\uu_2, \cond)$ etc.
Our CCB design is illustrated in \cref{fig:conditional-coupling}.
Multiple coupling blocks are then stacked to form the INN-part of the cINN.
We denote the entire INN as $f(x; c, \theta)$, with the network parameters $\theta$ 
and the inverse as $g(z; c, \theta)$.
Because the resolution does not stay fixed throughout the INN, 
different sections of the network require different conditions $c^{(k)}$.
We then use $c \coloneqq \{ c^{(k)} \}$ to denote the set of all the conditions at once.
For any fixed condition $\cond$, the invertibility is given as 
\begin{equation}
  f^{-1}(\cdot \, ; \cond, \theta) = g(\cdot \, ; \cond, \theta).
\end{equation}

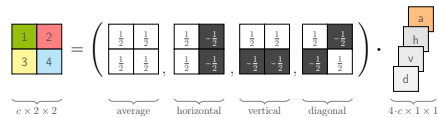
\begin{wrapfigure}{R}{0.5\linewidth}
    \vspace{-7mm}
        \resizebox{\linewidth}{!}{ \tikzstyle{box} = [rectangle, inner sep=1pt, outer sep=0pt, align=center, minimum width=1cm, minimum height=1cm, text depth=0, line width=1pt, draw=black, fill=white, font=\sffamily\Large, text=vll-dark]
\tikzstyle{dark} = [fill=vll-dark, text=white]
\tikzstyle{brace} = [decoration={brace, mirror, raise=0mm, amplitude=2mm}, decorate, black!50]
\tikzstyle{bracelabel} = [below=1mm, text=vll-dark, font=\large]

{\huge
\begin{math}
\begin{tikzpicture}[baseline=-0.65ex]
    \node [box, fill=vll-green] (1) at (-0.5, 0.5) {1};
    \node [box, fill=red!50] (2) at ( 0.5, 0.5) {2};
    \node [box, fill=yellow!50] (3) at (-0.5,-0.5) {3};
    \node [box, fill=cyan!25] (4) at ( 0.5,-0.5) {4};
    \draw [brace] (-1,-2) -- node [bracelabel] {$c \times 2 \times 2$} (1,-2);
\end{tikzpicture}\,
=
\Biggl(\;
\begin{tikzpicture}[baseline=-0.65ex]
    \node [box] (1) at (-0.5, 0.5) {$\frac{1}{2}$};
    \node [box] (2) at ( 0.5, 0.5) {$\frac{1}{2}$};
    \node [box] (3) at (-0.5,-0.5) {$\frac{1}{2}$};
    \node [box] (4) at ( 0.5,-0.5) {$\frac{1}{2}$};
    \draw [brace] (-1,-2) -- node [bracelabel] {average\vphantom{l}} (1,-2);
\end{tikzpicture}
\tikz[baseline=-0.65ex] \node[inner sep=0] at (0,-1) {\,,\;};
\begin{tikzpicture}[baseline=-0.65ex]
    \node [box] (1) at (-0.5, 0.5) {$\frac{1}{2}$};
    \node [box, dark] (2) at ( 0.5, 0.5) {$\sminus\frac{1}{2}$};
    \node [box] (3) at (-0.5,-0.5) {$\frac{1}{2}$};
    \node [box, dark] (4) at ( 0.5,-0.5) {$\sminus\frac{1}{2}$};
    \draw [brace] (-1,-2) -- node [bracelabel] {horizontal} (1,-2);
\end{tikzpicture}
\tikz[baseline=-0.65ex] \node[inner sep=0] at (0,-1) {\,,\;};
\begin{tikzpicture}[baseline=-0.65ex]
    \node [box] (1) at (-0.5, 0.5) {$\frac{1}{2}$};
    \node [box] (2) at ( 0.5, 0.5) {$\frac{1}{2}$};
    \node [box, dark] (3) at (-0.5,-0.5) {$\sminus\frac{1}{2}$};
    \node [box, dark] (4) at ( 0.5,-0.5) {$\sminus\frac{1}{2}$};
    \draw [brace] (-1,-2) -- node [bracelabel] {vertical} (1,-2);
\end{tikzpicture}
\tikz[baseline=-0.65ex] \node[inner sep=0] at (0,-1) {\,,\;};
\begin{tikzpicture}[baseline=-0.65ex]
    \node [box] (1) at (-0.5, 0.5) {$\frac{1}{2}$};
    \node [box, dark] (2) at ( 0.5, 0.5) {$\sminus\frac{1}{2}$};
    \node [box, dark] (3) at (-0.5,-0.5) {$\sminus\frac{1}{2}$};
    \node [box] (4) at ( 0.5,-0.5) {$\frac{1}{2}$};
    \draw [brace] (-1,-2) -- node [bracelabel] {diagonal} (1,-2);
\end{tikzpicture}
\;\Biggr)
\boldsymbol{\cdot}\!
\begin{tikzpicture}[baseline=-0.65ex]
    \node [box, fill=orange!50] (1) at ( 0.3, 1.2) {a};
    \node [box, fill=black!11] (2) at ( 0.1, 0.4) {h};
    \node [box, fill=black!8] (3) at (-0.1,-0.4) {v};
    \node [box, fill=black!5] (4) at (-0.3,-1.2) {d};
    \draw [brace] (-0.9,-2) -- node [bracelabel] {$4 \!\cdot\! c \times 1 \times 1$} (0.9,-2);
\end{tikzpicture}
\end{math}
} }%
        \vspace{-2mm}
        \captionof{figure}{Haar wavelet downsampling reduces spatial dimensions \& separates lower frequencies (a) from high (h,v,d).}
        \label{fig:haar_wavelets}
    \vspace{-6mm}
\end{wrapfigure}

\noindent \textbf{Haar wavelet downsampling.}
All prior INN architectures use one of two checkerboard patterns for reshaping to lower spatial resolutions (\cite{dinh2016density} or \cite{jacobsen2018irevnet}).
Instead, we find it helpful to perform downsampling with Haar wavelets \cite{haar1910wavelet},
which essentially decompose images into a $2\times 2$ average pooling channel 
as well as vertical, horizontal and diagonal derivatives, see \cref{fig:haar_wavelets}.
This results in a more sensible way of distributing the information after downsampling
and also contributes to mixing the variables between resolution levels.
Similarly, \cite{jacobsen2018excessive} use a single discrete cosine transform as a final transformation in their INN, to replace global average pooling.\\[2mm]
\noindent
\textbf{Conditioning network.}
It is the task of the conditioning network to transform the original condition $y$
into the necessary features $c^{(k)}$ that the INN uses at the different resolution levels $k$.
For this, we simply use a standard feed-forward network, denoted $\cnet$,
that jointly outputs the different features in the form of the feature pyramid $c$.
The conditioning network can be trained from scratch, jointly with the INN-part,
as explained in the next section.
It is also possible to use a pretrained model for initialization to speed up the start of training,
e.g. a pretrained ResNet \cite{he2016deep} or VGG \cite{simonyan14vgg}.


\subsection{Maximum likelihood training of cINNs}
\noindent
\textbf{Training the INN-part.}
By prescribing a probability distribution $p_Z(z)$ on latent space $z$, the INN $f$ assigns any input $x$ a probability, 
dependent on the conditioning $\cond$ and the network parameters $\theta$, through the change-of-variables formula:
\begin{equation}
  q(x \mid \cond, \theta) = p_Z\left( f(x; \cond, \theta) \right) 
  \left| \,\text{det}\!\left( \frac{\partial f}{\partial x}\right)\right| \ .
  \label{eq:change_of_variables}
\end{equation}
Here, we use the Jacobian matrix ${\partial f}/{\partial x}$. 
We will denote the Jacobian determinant, evaluated at some training sample $x_i$, as
$ J_i \coloneqq \text{det}\big( {\partial f}/{\partial x}|_{x_i} \big)$.
With a set of observerd i.i.d.~samples $\{ (x_i, c_i) \}$,
Bayes' theorem gives us the posterior over model parameters as 
\begin{equation}
p(\theta \mid \{(x_i, \cond_i) \}) \propto p_\theta(\theta) \prod_i q(\x_i \mid \cond_i, \theta) 
\label{eq:bayes_parameter_posterior}
\end{equation}
This means we can find the most likely model parameters given the known training data by maximizing the right hand side.
After taking the logarithm and changing the product to a sum, we get the following loss to minimize:
  $\mathcal{L} = \mathbb{E}_i\left[- \log\big(q(x_i \mid \cond_i, \theta)\big)\right]$,
which is the same as in classical Bayesian model fitting.
Finally, inserting \cref{eq:change_of_variables} with a standard normal distribution for $p_Z(z)$, we obtain
the {\it conditional maximum likelihood loss} we use for training:
\begin{equation}
  \cml = \mathbb{E}_i\!\left[\frac{\| f(\x_i; \cond_i, \theta)\|_2^2}{2} - \log\big| J_i \big| \right]. 
  \label{eq:ml_loss}
\end{equation}
We can also explicitly include a Gaussian prior over weights $p_\theta = \mathcal{N}(0, \sigma_\theta)$ in \cref{eq:bayes_parameter_posterior},
which amounts to the commonly used L2 weight regularization in practice.
Training a network with this loss yields an estimate of the maximum likelihood network parameters $\hat\theta$.
From there, we can perform conditional generation for some $\cond$ by sampling $\z$ and using the inverted network $g$:
$  \x_\mathrm{gen} = g(\z; \cond, \hat\theta)$, with $\z\sim p_Z(\z)$.

The maximum likelihood training method makes it virtually impossible for mode collapse to occur:
If any mode in the training set has low probability under the current guess $q(\x \mid \cond, \theta)$, 
the corresponding latent vectors will lie far outside the normal distribution $p_Z$ and receive big loss from the first L2-term in \cref{eq:ml_loss}.
In contrast, the discriminator of a GAN only supplies a weak signal, proportional to the mode's relative frequency in the training data,
so that the generator is not penalized much for ignoring a mode completely.\\[3mm]
\noindent
\textbf{Jointly training the conditioning network.}
Next, we consider the result if we also backpropagate the loss through the feature pyramid $c$, to train the conditioning network $\cnet$ jointly with the same loss.
Intuitively speaking, the more useful the learned features are for the INN's task,
the lower the $\cml$ loss will become. 
Therefore, the conditioning network is encouraged to extract useful features.

We can formalize this using the information-theoretical concept of mutual information (MI).
MI quantifies the amount of information that two variables share, in other words, how informative one variable is about the other.
For any two random variables $a$ and $b$, It can be written as the KL-divergence between joint and factored distributions:
$I(a,b) = D_\mathrm{KL}(p(a,b) \| p(a)p(b))$.
With this, we can derive the following proposition, details and proof are found in the appendix: \\ \noindent
\textbf{Proposition 1. }{\it
Let $\hat \theta$ be the INN parameters and $\hat \cnet$ the conditioning network that jointly minimize $\cml$.
Assume that the INN $f(\cdot; \cdot, \theta)$ is optimized over $\mathcal{F}$ defined in Assumption 1 (appendix),
and $\cnet$ over $\mathcal{G}_0$ defined in Assumption 2 (appendix).
Then it holds that
\begin{equation}
    I \big(x, \hat \cnet(y) \big) = \underset{\varphi \in \mathcal{G}_0}{\operatorname {max}} \; I \big(x, \cnet(y) \big) 
\end{equation}
}
In other words, the learned features will be the ones that are maximally informative about the generated variable $x$.
Importantly, the assumption about the conditioning networks family $\mathcal{G}_0$ does not say anything about its representational power:
the features will be as informative as possible within the limitations of the conditioning network's architecture and number of extracted features.

We can go a step further under the assumption that the power of the conditioning network and number of features in the pyramid
are large enough to reach the global minimum of the loss (sufficient condition given by Assumption 3, appendix).
In this case, we can also show that the cINN as a whole will learn the true posterior by minimizing the loss (proof in appendix): \\ \noindent
\textbf{Proposition 2. }{\it
Assume $\cnet$ has been optimized over a family $\mathcal{G}_1$ of universal approximators 
and $\dim(c) \geq \dim(y)$ (Assumption 3, appendix),
and the INN is optimized over a family of universal density approximators $\mathcal{F}$ (Assumption 1, appendix).
Then the following holds for $(x,y) \in \mathcal{X}$, where $\mathcal{X}$ is the joint domain of the true training distribution $p(x,y)$:
\begin{equation}
    q(x | \hat \cnet(y), \hat \theta) = p(x|y)
\end{equation}
}
\vspace{-12mm}

\section{Experiments}
\label{sec:experiments}
We present results and explore the latent space of our models for two image-to-image generation tasks: 
day to night image translation, and image colorization.
We use the former as a qualitative demonstration, and the latter for a more in-depth analysis and comparison with other methods.
MNIST experiments, to purely show the capability of the CCBs without the conditioning network,
are given in the appendix.

\begin{figure}
    \input{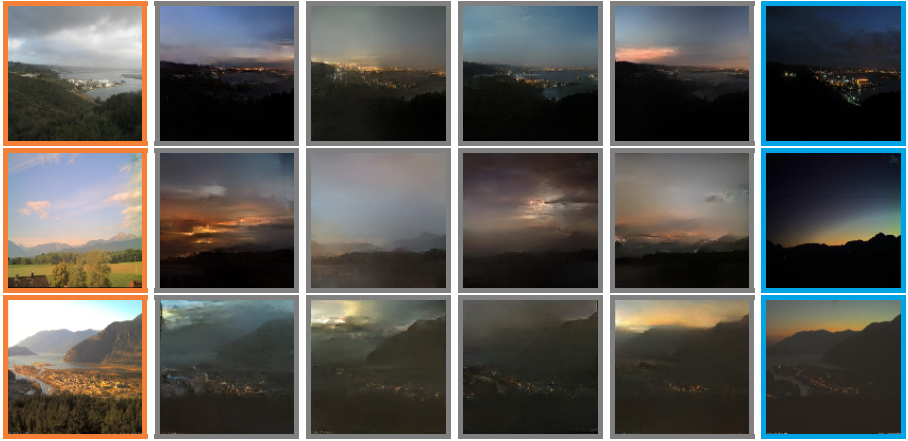}
    \vspace{-4mm}
    \caption{
    Examples of \textcolor{Orange}{\bf conditions $y$} (left), 
    three \textcolor{black!60}{\bf generated samples} (middle), 
    and the \textcolor{Cerulean}{\bf original image $x$} (right).
    }
    \vspace{-4mm}
    \label{fig:maps_samples}
\end{figure}

In practice, we use several techniques to improve the network and training. 
Ablations of the following are included in the appendix.
\begin{itemize}
\item We augment the images by adding a small amount of noise, in order to remove the quanitzation into 255 brightness levels.
The quantization is known to cause problems in training otherwise \cite{theis2015note}.
\item After each coupling block, we perform a random, fixed permuation of the feature channels.
    This effectively randomizes the split for the next coupling block.
\item We adopt the method from \cite{dinh2016density}, whereby the affine scaling $s$
    is parametrized as $\gamma \, \mathrm{tanh}(r(x))$, where $\gamma$ is learned directly as a channel-wise
    parameter, and $r$ is output by the subnetwork.
    This has exactly the same representational power as directly outputting $s$, but improves stability,
    because the $\exp(s)$ term in \cref{eq:unconditional_forward} does not explode as easily.
\end{itemize}

\subsection{Day to Night Translation}

\begin{wrapfigure}{r}{0.45\textwidth}
\vspace{-10mm}
    \centering
        \input{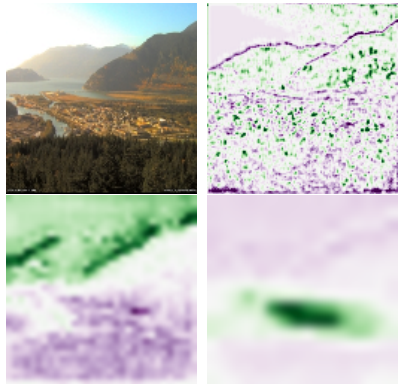}
        \vspace{-4mm}
        \captionof{figure}{
        Conditioning image (top left),
        and extracted features from different levels of the pyramid.
        From left to right, top to bottom: 
        1st level, precise edges and texture; 
        2nd level, foreground/background; 
        3rd level, populated area.}
        \label{fig:pca_feature_pyramid}
    \vspace{-9mm}
\end{wrapfigure}

We train on the popular day-to-night dataset \cite{laffont2014transient}.
It contains webcam images from approximately 100 different locations,
taken at approximately 10-20 times during the day and night each.
This results in about $200$ combinations of day-night pairs per location.
The test set consists of 5 unseen locations.
For training, we randomly resize and crop the images to $128 \times 128$ pixels.
We use the day-images as the condition $y$, and the night-images as the generated $x$.
For the conditioning network, we use a standard ResNet-18 \cite{he2016deep}. 
We extract the activations after every other layer of the ResNet to form the feature pyramid.
As the ResNet contains the usual downsampling operations, the activations already have the correct sizes for the pyramid.
We then construct the INN part as described in Sec.~\ref{sec:methods},
with 8 coupling blocks in total, and five wavelet downsampling operations spaced in between.
The subnetworks consist of three convolutions, with ReLU activations and batch normalization after the first two convolutions.

We train for $175\,000$ iterations using the Adam optimizer, with a batch-size of 48,
and leave the learning rate fixed at 0.001 throughout.
These training parameters are comparable to those of standard feed-forward models.

Despite the relatively small training set, we see little signs of overfitting, and the model generalizes well to the test set.
Previously, \cite{sun2019dual} also found low overfitting and good generalization on small training sets using INNs.
Several samples by the model are shown in Fig.~\ref{fig:maps_samples}. The cINN correctly recognizes populated regions and generates lights there,
as well as freely synthesizing diverse cloud patterns and weather conditions.
At the same time, the edges and structures (e.g. mountains) are correctly aligned with the conditioning image.
The features learned by the conditioning network are visualized in Fig.~\ref{fig:pca_feature_pyramid}.
Hereby, independent features were extracted via PCA. The figure shows one example of a feature from the first three levels of the pyramid.


\subsection{Diverse image colorization}
\label{sec:colorization}
\noindent
For a more challenging task, we turn to colorization of natural images.
The common approach for this task is to represent images in $Lab$ color space and generate color channels $\bolda, \boldb$ by a model conditioned on the luminance channel $\boldL$.
We train on the ImageNet dataset \cite{russakovsky15imagenet}.
As the color channels do not require as much resolution as the luminance channel, 
we condition on $256\times256$ pixel grayscale images, but generate $64 \times 64$ pixel color information.
This is in accordance with the majority of existing colorization methods.

\begin{figure}[!h!]

    \parbox{0.48\linewidth}{
	\parbox{1.0\linewidth}{
    \includegraphics[width=0.95\linewidth]{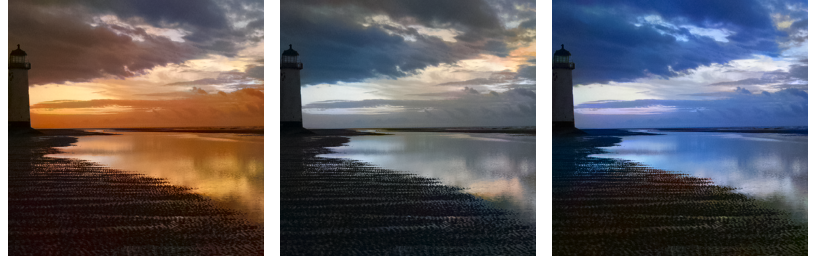} \\[1.2mm]
    \includegraphics[width=0.95\linewidth]{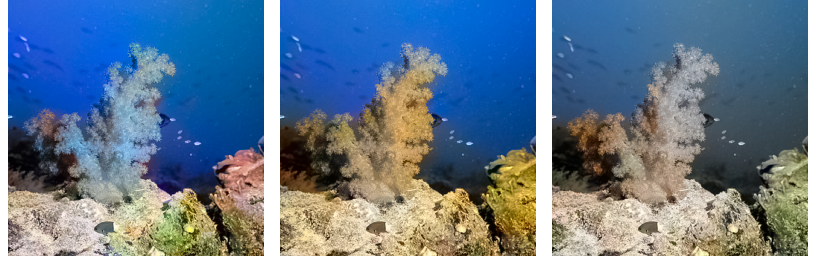} \\[1.2mm]
    \includegraphics[width=0.95\linewidth]{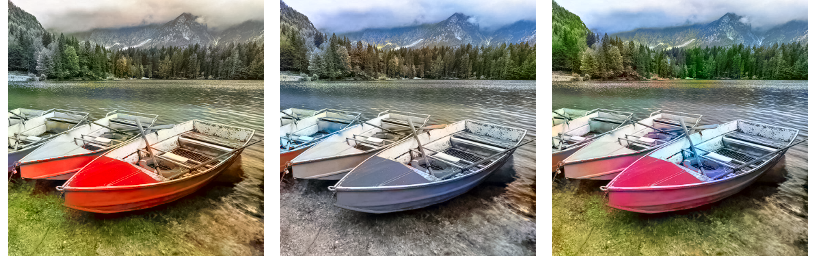} \\[1.2mm]
    \includegraphics[width=0.95\linewidth]{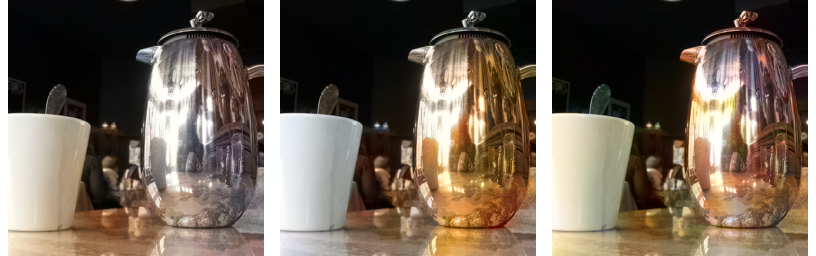} \\[1.2mm]
    \includegraphics[width=0.95\linewidth]{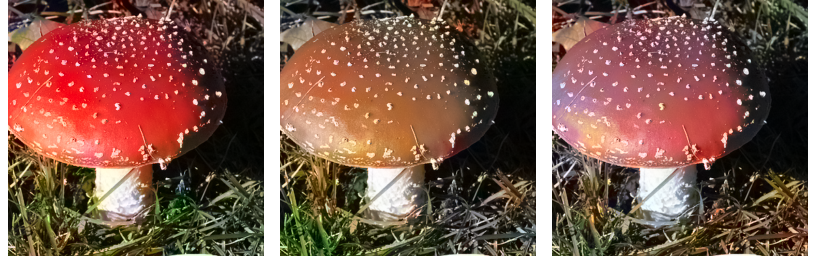} \\[1.2mm]
    \includegraphics[width=0.95\linewidth]{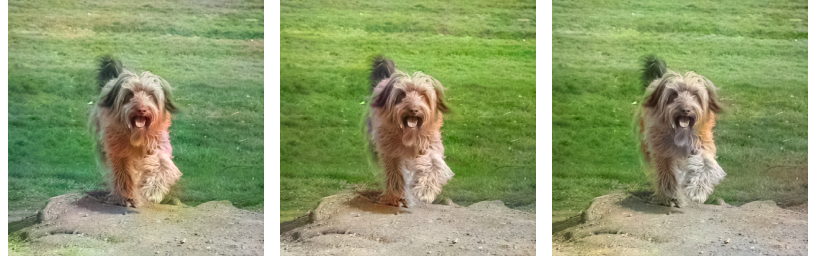}
    \vspace{-1mm}
	\caption{Diverse colorizations produced by our cINN.}
	\label{fig:colorization_examples}}}
	\quad
    \parbox{0.48\linewidth}{
    \parbox{\linewidth}{
        \centering
        \parbox{0.93\linewidth}{
        \includegraphics[width=\linewidth]{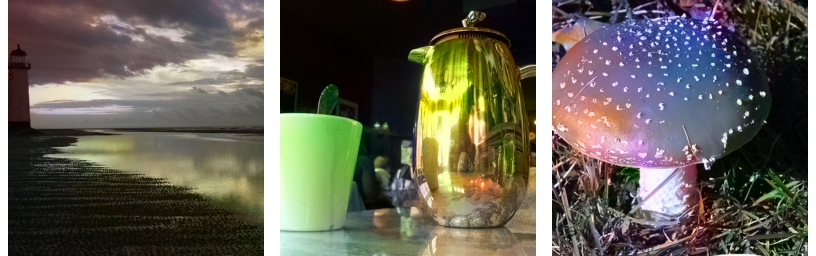}\\[1.0mm]
        \includegraphics[width=\linewidth]{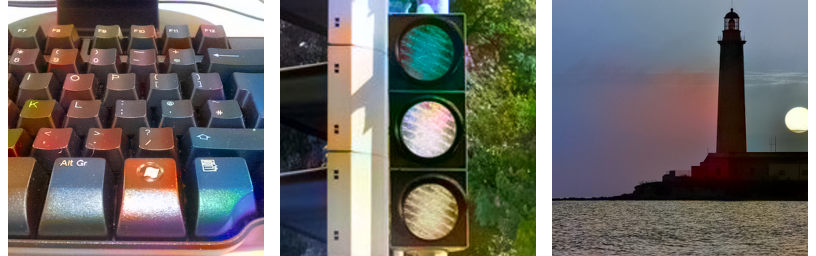} 
        }
        \vspace{-2mm} 
    	\caption{Failure cases of our method.
    	{\em Top:} Sampling outliers.
    	{\em Bottom:} cINN did not recognize an object's semantic class or connectivity.
    	}
    	\label{fig:failure_cases}}
    	
    \parbox{\linewidth}{
        \centering
    	{ \def\svgwidth{0.93\linewidth} 
\begingroup%
  \makeatletter%
  \providecommand\color[2][]{%
    \errmessage{(Inkscape) Color is used for the text in Inkscape, but the package 'color.sty' is not loaded}%
    \renewcommand\color[2][]{}%
  }%
  \providecommand\transparent[1]{%
    \errmessage{(Inkscape) Transparency is used (non-zero) for the text in Inkscape, but the package 'transparent.sty' is not loaded}%
    \renewcommand\transparent[1]{}%
  }%
  \providecommand\rotatebox[2]{#2}%
  \newcommand*\fsize{\dimexpr\f@size pt\relax}%
  \newcommand*\lineheight[1]{\fontsize{\fsize}{#1\fsize}\selectfont}%
  \ifx\svgwidth\undefined%
    \setlength{\unitlength}{405.38953886bp}%
    \ifx\svgscale\undefined%
      \relax%
    \else%
      \setlength{\unitlength}{\unitlength * \real{\svgscale}}%
    \fi%
  \else%
    \setlength{\unitlength}{\svgwidth}%
  \fi%
  \global\let\svgwidth\undefined%
  \global\let\svgscale\undefined%
  \makeatother%
  \begin{picture}(1,1.23500303)%
    \lineheight{1}%
    \setlength\tabcolsep{0pt}%
    \put(0,0){\includegraphics[width=\unitlength,page=1]{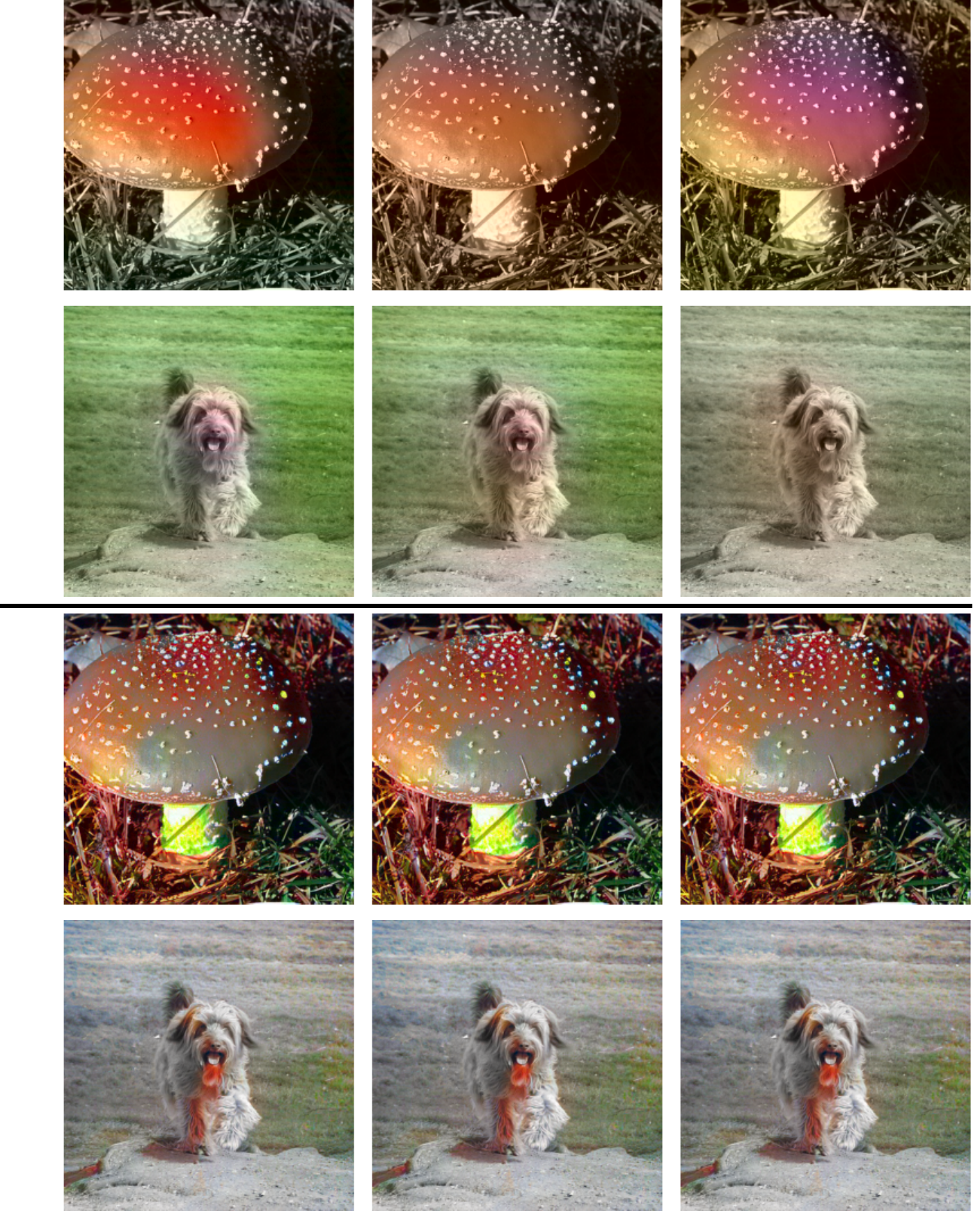}}%
    \put(0.05840434,0.89083765){\color[rgb]{0,0,0}\rotatebox{90}{\makebox(0,0)[lt]{\lineheight{1.25}\smash{\begin{tabular}[t]{l}VAE\end{tabular}}}}}%
    \put(0.05840434,0.24035675){\color[rgb]{0,0,0}\rotatebox{90}{\makebox(0,0)[lt]{\lineheight{1.25}\smash{\begin{tabular}[t]{l}cGAN\end{tabular}}}}}%
  \end{picture}%
\endgroup%
}
        \vspace{-3mm}
    	\caption{
    	Other methods have lower diversity or quality, 
        and suffer from inconsistencies in objects, or color blurriness and bleeding
        (cf. \cref{fig:colorization_examples}, bottom).
    	}
    	\label{fig:colorization_comparison}}}
    	
	\parbox{\linewidth}{
	\centering
    \includegraphics[width=0.48\linewidth]{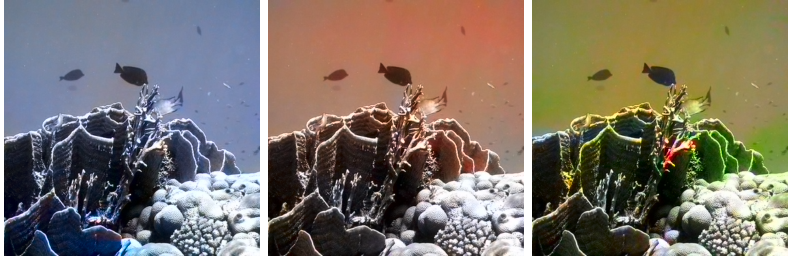}\hspace{2mm}%
    \includegraphics[width=0.48\linewidth]{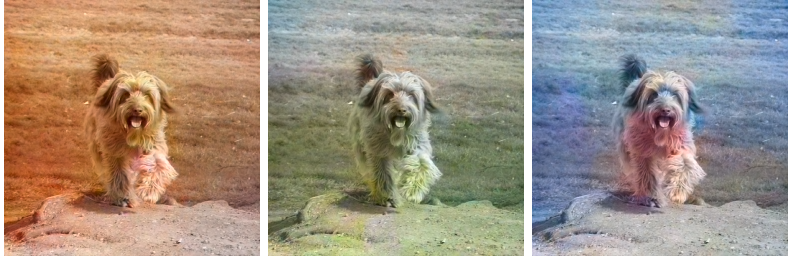}
    \vspace{-1mm}
	\caption{
	In an ablation study, we train a cINN using the grayscale image directly as conditional input, without a conditioning network $\cnet$.
	The resulting colorizations largely ignore semantic content which leads to exaggerated diversity.
	More ablations are found in the appendix.
    }
	\label{fig:ablation}}
\end{figure}

For the conditioning network $\cnet$, we start with the same VGG-like architecture from \cite{zhang2016colorful} and {pretrain on the colorization task} using their code.
We then cut off the network before the second-to-last convolution, 
resulting in 256 feature maps of size $64\times64$ from the grayscale image $\boldL$.
To form the feature pyramid, we then add a series of strided convolutions, ReLUs, and batch normaliziation layers on top,
to produce the features at each resolution.
The ablation study in \cref{fig:ablation} confirms that the conditioning network is 
absolutely necessary to capture semantic information.

The INN-part constist of 22 convolutional CCBs, with three downsampling steps in between.
After that, the features are flattened, followed by 8 fully connected CCBs.
To conserve memory and computation, we adopt a similar splitting- and merging-scheme as in \cite{dinh2014nice}:
after each wavelet downsampling step, we split off half the channels.
These are not processed any further, but fed into a skip connection and concatenated directly onto the latent output vector.
This way, the INN as a whole stays invertible.
The reasoning behind this is the following:
The high resolution stages have a smaller receptive field and less expressive power, 
so the channels split off early correspond to local structures and noise.
More global information is passed on to the lower resolution sections of the INN and processed further.
Overall, the generative performance of the network is not meaningfully impacted, 
while dramatically reducing the computational cost.

For training, we use the Adam optimizer for faster convergence, and train for roughly $250\,000$ iterations,
and a batch-size of 48.
The learning rate is $10^{-3}$, decreasing by a factor of 10 at $100\,000$ and $200\,000$ iterations.
At inference time, we use joint bilateral upsampling \cite{kopf2007joint} to match the resolution of the generated color channels ${\bolda}$, ${\boldb}$
to that of the luminance channel $\boldL$.
This produces visually slightly more pleasing edges than bicubic upsampling, but has little to no impact on the results.
It was not used in the quantitative results table, to ensure an unbiased comparison.

{Latent space interpolations and color transfer are shown in \cref{fig:latent_temperature,fig:latent_transfer},
with more experiments in the appendix.
In \cref{tab:results}, a quantitative comparison to existing methods is given.
The cINN clearly has the best sample diversity, as summarized by 
the variance and best-of-8 accuracy.
The standard cGAN 
\parfillskip=0pt\par}

\begin{figure}[h]
    \vspace{-3mm}
    \centering
	{ \def\svgwidth{\textwidth} 
\begingroup%
  \makeatletter%
  \providecommand\color[2][]{%
    \errmessage{(Inkscape) Color is used for the text in Inkscape, but the package 'color.sty' is not loaded}%
    \renewcommand\color[2][]{}%
  }%
  \providecommand\transparent[1]{%
    \errmessage{(Inkscape) Transparency is used (non-zero) for the text in Inkscape, but the package 'transparent.sty' is not loaded}%
    \renewcommand\transparent[1]{}%
  }%
  \providecommand\rotatebox[2]{#2}%
  \newcommand*\fsize{\dimexpr\f@size pt\relax}%
  \newcommand*\lineheight[1]{\fontsize{\fsize}{#1\fsize}\selectfont}%
  \ifx\svgwidth\undefined%
    \setlength{\unitlength}{392.07207627bp}%
    \ifx\svgscale\undefined%
      \relax%
    \else%
      \setlength{\unitlength}{\unitlength * \real{\svgscale}}%
    \fi%
  \else%
    \setlength{\unitlength}{\svgwidth}%
  \fi%
  \global\let\svgwidth\undefined%
  \global\let\svgscale\undefined%
  \makeatother%
  \begin{picture}(1,0.18382352)%
    \lineheight{1}%
    \setlength\tabcolsep{0pt}%
    \put(0,0){\includegraphics[width=\unitlength,page=1]{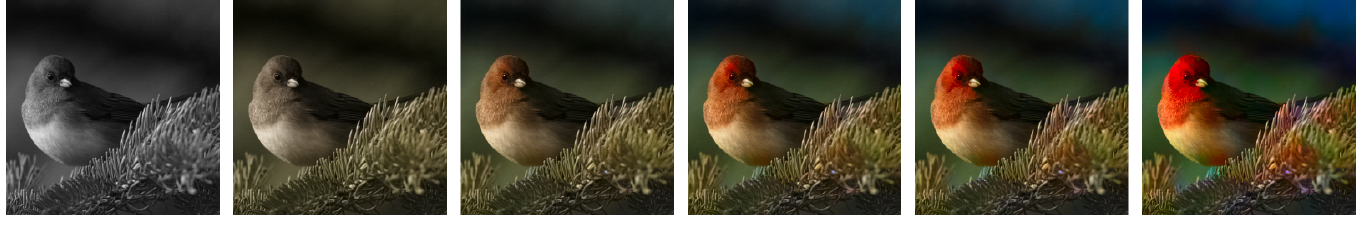}}%
    \put(0.00386467,0.00628917){\color[rgb]{0,0,0}\makebox(0,0)[lt]{\lineheight{1.25}\smash{\begin{tabular}[t]{l}Grayscale $y$\end{tabular}}}}%
    \put(0.17034886,0.00628017){\color[rgb]{0,0,0}\makebox(0,0)[lt]{\lineheight{1.25}\smash{\begin{tabular}[t]{l}$\z = 0.0 \cdot \z^*$\end{tabular}}}}%
    \put(0.337107,0.00628017){\color[rgb]{0,0,0}\makebox(0,0)[lt]{\lineheight{1.25}\smash{\begin{tabular}[t]{l}$\z = 0.7 \cdot \z^*$\end{tabular}}}}%
    \put(0.50386527,0.00628017){\color[rgb]{0,0,0}\makebox(0,0)[lt]{\lineheight{1.25}\smash{\begin{tabular}[t]{l}$\z = 0.9 \cdot \z^*$\end{tabular}}}}%
    \put(0.67062337,0.00628017){\color[rgb]{0,0,0}\makebox(0,0)[lt]{\lineheight{1.25}\smash{\begin{tabular}[t]{l}$\z = 1.0 \cdot \z^*$\end{tabular}}}}%
    \put(0.83738153,0.00628017){\color[rgb]{0,0,0}\makebox(0,0)[lt]{\lineheight{1.25}\smash{\begin{tabular}[t]{l}$\z = 1.25 \cdot \z^*$\end{tabular}}}}%
  \end{picture}%
\endgroup%
 }
    \includegraphics[width=\textwidth]{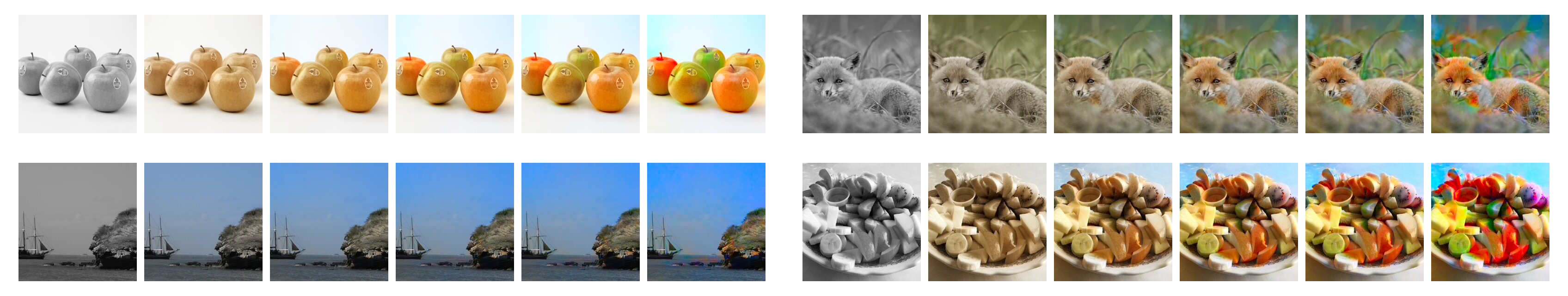}
    \vspace{-5mm}
	\captionof{figure}{Effects of linearly scaling the latent code $\z$ while keeping the condition fixed.
	Vector $\z^*$ is ``typical'' in the sense that $\|\z^*\|^2=\mathbb{E}\big[\|\z\|^2\big]$, and results in natural colors.
	As we move closer to the center of the latent space ($\|\z\| < \|\z^*\|$), regions with ambiguous colors become desaturated, while less ambiguous regions (e.g.~sky, vegetation) revert to their prototypical colors.
	In the opposite direction ($\|\z\| > \|\z^*\|$), colors are enhanced to the point of oversaturation.
	\vspace{1mm}}
	\label{fig:latent_temperature}
	
    \includegraphics[width=\textwidth]{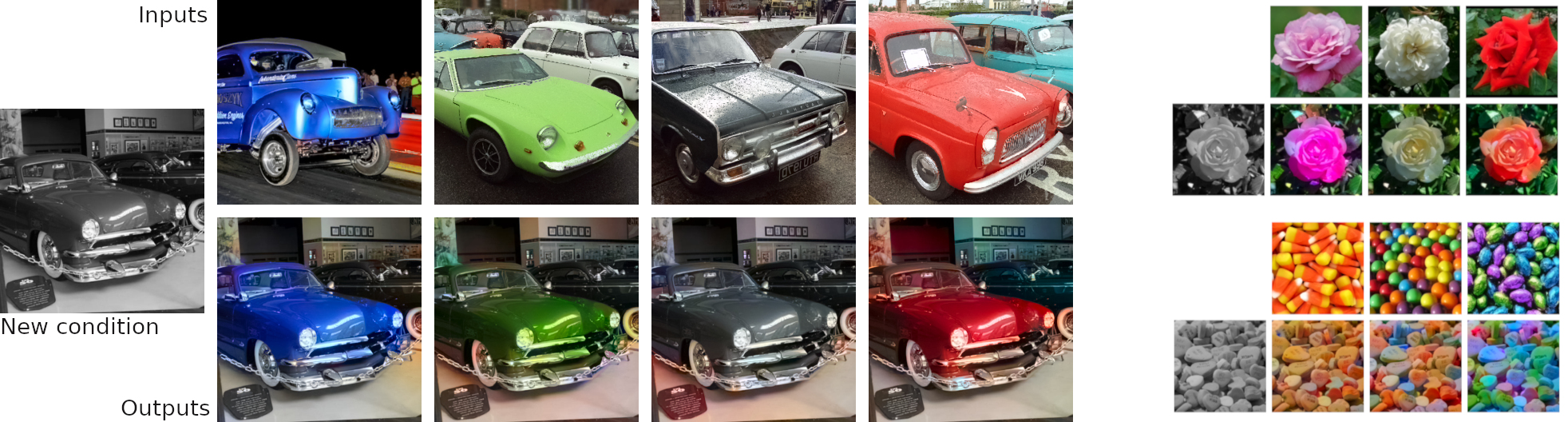}
    \vspace{-5mm}
	\captionof{figure}{For color transfer, we first compute the latent vectors $\z$ for different color images $(\boldL,\bolda,\boldb)$ \emph{(top row)}. 
	We then send the same $\z$ vectors through the inverse network with a new grayscale condition $\boldL^*$ \emph{(far left)}
	to produce transferred colorizations $\bolda^*,\boldb^*$ \emph{(bottom row)}.
	Differences between reference and output color (e.g.~pink rose) can arise from mismatches between the reference colors $\bolda,\boldb$ and the intensity prescribed by the new condition $\boldL^*$.
	\vspace{-5mm}
	}
	\label{fig:latent_transfer}
\end{figure}

%



\newpage

\begin{table*}[t!]
\centering
\resizebox{\linewidth}{!}{
\begin{tabular}{| l | r r r | r | r  r |}
\hline 
& cINN (ours)
& VAE-MDN
& cGAN
& CNN
& BW
& Ground truth \\
\hline 
MSE best of 8  
& {\bftab 3.53$\pm$0.04} 
&4.06$\pm$0.04  
& 9.75$\pm$0.06 
& 6.77 $\pm$0.05 
&  -- 
&  --  \\
Variance        
& {\bftab 35.2$\pm$0.3} 
& 21.1$\pm$0.2   
& 0.0$\pm$0.0 
&   --          
&   --          
&   --      \\
FID
& 25.13$\pm$0.30              
& 25.98$\pm$0.28         
& {\bftab 24.41$\pm$0.27}
& 24.95$\pm$0.27        
& $30.91\pm0.27$
&  14.69 $\pm$ 0.18 \\
VGG top 5 acc. 
& 85.00$\pm$0.48  
& 85.00$\pm$0.48 
& 84.62$\pm$0.53  
& {\bftab 86.86$\pm$0.41} 
& 86.02$\pm$0.43
& 91.66 $\pm$ 0.43\\
\hline 
\end{tabular}
}
\vspace{2mm}
\caption{Comparison of conditional generative models for diverse colorization
(VAE-MDN: \cite{deshpande2017learning}; 
cGAN: \cite{isola2017image}).
We additionally compare to a state-of-the-art regression model (`CNN', no diversity, \cite{iizuka2016let}),
and the grayscale images alone (`BW').
For each of 5k ImageNet validation images,
we compare the best pixel-wise MSE of 8 generated colorization samples, 
the pixel-wise variance between the 8 samples as an approximation of the diversity,
the Fréchet Inception Distance \cite{heusel2017gans} as a measure of realism,
and the top 5 accuracy of ImageNet classification performed on the colorized images, 
to check if semantic content is preserved by the colorization. 
\vspace{-6mm}}
\label{tab:results}
\end{table*}

\newpage
\noindent
completely ignores the latent code, and relies only on the condition.
As a result, we do not observe any measurable diversity, in line with results from \cite{isola2017image}.
In terms of FID score, the cGAN performs best, 
although its results do not appear more realistic to the human eye, cf.~\cref{fig:colorization_comparison}.
This may be due to the fact that FID is sensitive to outliers, which are unavoidable for a truly diverse method (see \cref{fig:failure_cases}), 
or because the discriminator loss implicitly optimizes for the similarity of deep CNN activations.
{The VGG classification accuracy of colorized images is decreased for all generative methods equally,
because occasional outliers may lead to misclassification.}

\section{Conclusion and Outlook}
\noindent We have proposed a conditional invertible neural network architecture which enables 
diverse image-to-image translation with high realism.
For image colorization, we believe that even better results can be achieved when employing the latest tricks from large-scale GAN frameworks.
Especially the non-invertible nature of the conditioning network makes cINNs a suitable method for other computer vision tasks such as diverse semantic segmentation.

\section*{Acknowledgements}
This work is supported by Deutsche Forschungsgemeinschaft (DFG) under Germany’s Excellence Strategy EXC-2181/1 - 390900948 (the Heidelberg STRUCTURES Excellence Cluster). 
LA received funding by the Federal Ministry of Education and Research of Germany project High Performance Deep Learning Framework (No 01IH17002).
JK was supported by by Informatics for Life funded by the Klaus Tschira Foundation. 
CR and UK received financial support from the European Re-search Council (ERC) under the European Unions Horizon2020 research and innovation program (grant agreement No647769).

\bibliographystyle{splncs04}
\bibliography{bibliography}

\end{document}